\documentclass[journal]{IEEEtran}

\usepackage{amsmath, amssymb, amsfonts}
\usepackage{graphicx} 
\usepackage{subcaption} 
\usepackage{tcolorbox}
\tcbuselibrary{skins}

\usepackage{pgfplots}
\pgfplotsset{compat=1.17} 

\usepackage{algorithmic}
\usepackage{algorithm}
\usepackage{cite}
\usepackage{stfloats}
\usepackage{textcomp}
\usepackage{url}

\usepackage{multirow}
\usepackage{array}
\usepackage{tabularx}
\usepackage{booktabs}
\usepackage{adjustbox}
\usepackage{makecell}
\usepackage[T1]{fontenc}

\usepackage[caption=false,font=normalsize,labelfont=sf,textfont=sf]{subfig}

\usepackage{verbatim}

\title{SPECTRUM: Semantic Processing and Emotion-informed video-Captioning Through Retrieval and Understanding Modalities}

\author{%
	Ehsan Faghihi, 
	Mohammedreza Zarenejad, and 
	Ali-Asghar Beheshti Shirazi%
	\thanks{Ehsan Faghihi is with the Department of Electrical Engineering, Iran University of Science and Technology, Tehran, IRAN (e-mail: ehsanfgh.commeng@gmail.com).}
	\thanks{Mohammedreza Zarenejad is with the Department of Electrical Engineering, Iran University of Science and Technology, Tehran, IRAN (e-mail: mr.zarenejad1@gmail.com).}
	\thanks{Ali-Asghar Beheshti Shirazi is with the Department of Electrical Engineering, Iran University of Science and Technology, Tehran, IRAN (e-mail: abeheshti@iust.ac.ir).}
}

\begin{document}
	
	\maketitle
	
	\IEEEpubidadjcol
	
	\begin{abstract}
     Capturing a video's meaning and critical concepts by analyzing the subtle details is a fundamental yet challenging task in video captioning. Identifying the dominant emotional tone in a video significantly enhances the perception of its context. Despite a strong emphasis on video captioning, existing models often need to adequately address emotional themes, resulting in suboptimal captioning results. To address these limitations, this paper proposes a novel Semantic Processing and Emotion-informed video-Captioning Through Retrieval and Understanding Modalities (SPECTRUM) framework to empower the generation of emotionally and semantically credible captions. Leveraging our pioneering structure, SPECTRUM discerns multimodal semantics and emotional themes using Visual-Text Attribute Investigation (VTAI) and determines the orientation of descriptive captions through a Holistic Concept-Oriented Theme (HCOT), expressing emotionally-informed and field-acquainted references. They exploit video-to-text retrieval capabilities and the multifaceted nature of video content to estimate the emotional probabilities of candidate captions. Then, the dominant theme of the video is determined by appropriately weighting embedded attribute vectors and applying coarse- and fine-grained emotional concepts, which define the video's contextual alignment. Furthermore, using two loss functions, SPECTRUM is optimized to integrate emotional information and minimize prediction errors. Extensive experiments on the EmVidCap, MSVD, and MSR-VTT video captioning datasets demonstrate that our model significantly surpasses state-of-the-art methods. Quantitative and qualitative evaluations highlight the model's ability to accurately capture and convey video emotions and multimodal attributes.


\end{abstract}							
\section{INTRODUCTION}		
\begin{figure}[!t]
	\centering
	\includegraphics[height = 7.6cm, width=9.0cm]{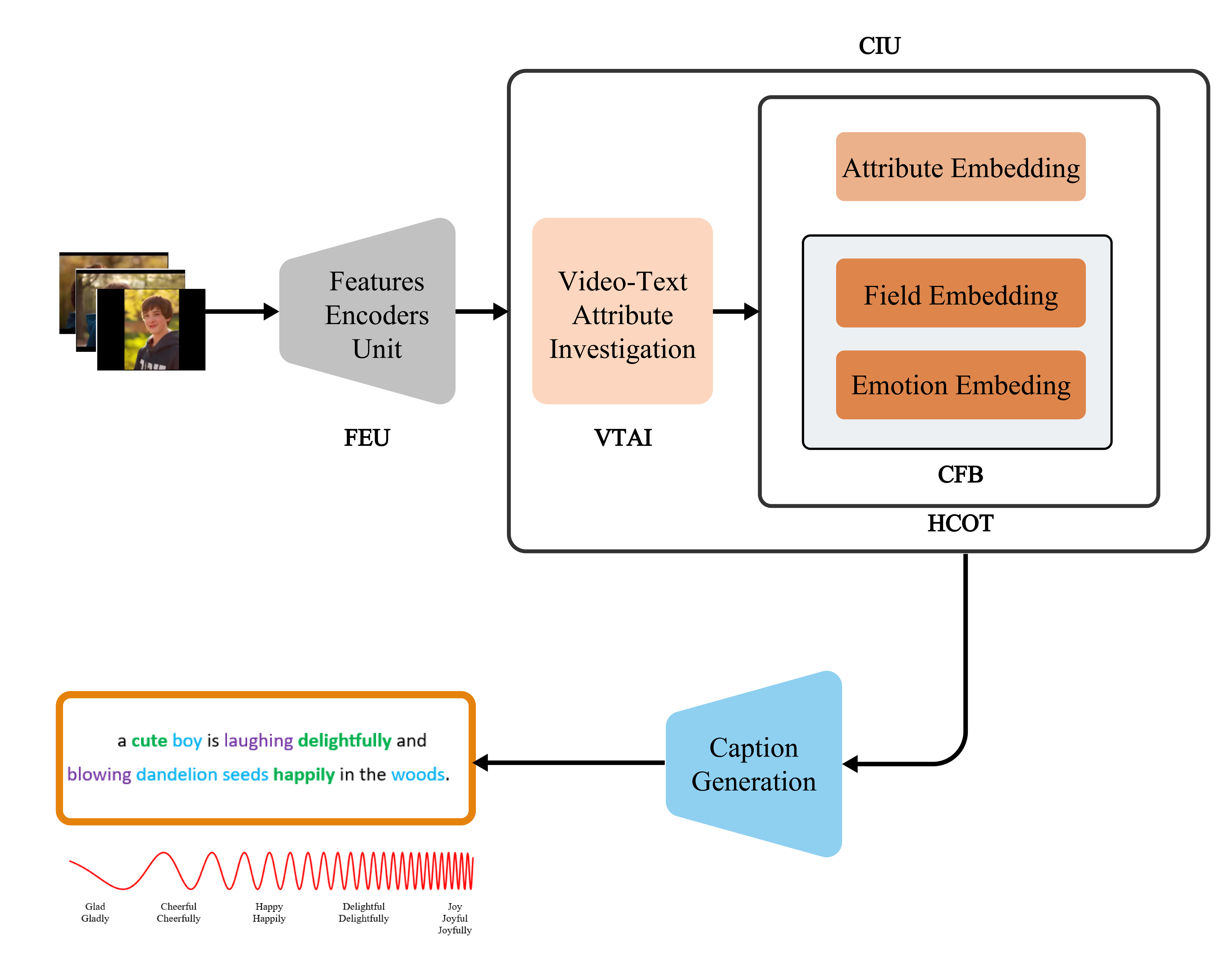}
	\caption{\footnotesize Diagram of the SPECTRUM framework illustrating the caption generation procedure of a query video based on the emotional spectrum. The SPECTRUM architecture is designed to generate emotionally-informed and factually-acquainted captions through the use of a concept investigation unit (CIU), including video-text attribute investigation (VTAI) and holistic concept-oriented theme (HCOT).}
	\label{fig1_structure}
	\vspace{-2em}
\end{figure}

 \IEEEPARstart{R}{ecent} advancements in computer vision and natural language processing have led to emerging joint vision-language research domains. Semantically describing videos represents a significant interdisciplinary challenge within these domains, aiming to generate precise textual descriptions from visual features automatically. Achieving a meaningful understanding of video content and translating it into precise and concise corresponding sentences constitutes the most critical challenge in video captioning. Considering the diverse underlying spatial and temporal relationships between visual features and their corresponding semantic connections is crucial to resolving this issue,\cite{1_9741388}, \cite{3_9762283}, \cite{4_9367203}. Video subtitling, video summarization, content-based video retrieval, human-computer interaction, assistance for the visually impaired, and video surveillance represent just a few of the myriad applications of semantic video description, \cite{8_9063637}, \cite{10_9380441}, \cite{11_article}.\\ 
Emotions and motions significantly enhance the richness of video captions due to their strong interconnection \cite{105_liu2020sentiment}. Researchers increasingly use emotion detection in natural language processing (NLP) to understand emotions in various contexts, classifying and measuring emotional intensities \cite{106_Ekman1992-EKMAAF}, \cite{107_PLUTCHIK19803}.  Effective video emotional captioning aims to improve sentence expressiveness by incorporating multimodal analysis—visual, audio, and textual information—thereby completely understanding a video's thematic and emotional contexts.\\
This paper introduces procedures designed to enhance vision-language correspondence and derive emotionally-informed and field-acquainted concepts, focusing on multimodal semantic analysis. The primary challenges associated with these procedures include the accurate extraction of attribute features and their effective utilization. Previous research has often failed to simultaneously consider both emotional and factual concepts in video description, highlighting a clear gap in the existing literature, \cite{21_9738841}, \cite{22_9864282}; our model is designed to be emotionally-informed and field-acquainted, utilizing a fine-to-coarse concept investigation approach to perform multimodal semantic analysis. Attribute features encompass both emotional and factual features, where factual words typically include concepts such as  \textit{beings}, \textit{objects}, \textit{scenes}, and most importantly, \textit{motions}. Emotional concepts can be categorized into 34 distinct bags of emotions, \cite{107_PLUTCHIK19803}, \cite{30_9352546}. Each 'bag of emotions,' considered a coarse emotional category, comprises a set of finer emotional terms that convey the same concept. For example, the "\textit{Calm}" category, a bag of emotions, includes finer terms such as "\textit{peaceful}," "\textit{quiet}," "\textit{calmly}," "\textit{quietly}," and "\textit{peacefully}" all of which reflect the same underlying concept as the coarse term. Hence, for each query video, a holistic theme oriented around both emotional and factual concepts predominates in the generated captions.\\
Consider Fig \ref{fig1_structure} as an example; our proposed model aims to identify the emotions, actions, and associated descriptive fields that comprehensively describe the video, thereby conceptually guiding its orientation. it becomes challenging to specify the emotion associated with \textit{playing music} without employing a concept-oriented theme that is holistically emotional. This is because the motion may arise from a range of emotions such as \textit{joy}, \textit{happiness}, \textit{sadness}, \textit{calmness}, \textit{anger}, \textit{frustration}, \textit{surprise}, or \textit{excitement}.		
In most mainstream studies, the coarse-to-fine investigation of attribute features often overlooks the combined impact of emotional and factual concepts, resulting in incomplete and partial captions. 
In this paper, a novel Semantic Processing and Emotion-informed video-Captioning Through Retrieval and Understanding Modalities (SPECTRUM) framework is designed to support the generation of credible semantic descriptions of videos, focusing on the dominant emotion associated with each motion. Moreover, we enrich our proposed architecture by incorporating the contrastive vision-language pre-trained model, CLIP  \cite{26_radford2021learning}, for encoding and video-to-text retrieval procedures.
In the Concept Investigation Unit (CIU) of the proposed architecture, the Video-Text Attribute Investigation (VTAI) stage conducts multimodal semantic analysis. It evaluates the likelihood of attribute concepts appearing in ground truth captions. In the second stage, the Holistic Concept-Oriented Theme (HCOT) aims to define the conceptual orientation of captions, reflecting emotionally-informed and field-acquainted sentences. This stage leverages the Coarse-to-Fine Block (CFB) and the Attribute Embedding Block (AEB) to prevent the emergence of semantically unrelated descriptions. By utilizing both textual and video features as inputs, the CIU enhances the model's robustness against ambiguities and inaccuracies during the semantic description of the video.\\
Our contributions to this work are as follows:\\
\textbf{I:} We present SPECTRUM, a novel framework for video captioning that incorporates the Concept Investigation Unit (CIU) as its core component. SPECTRUM enhances the generation of accurate semantic video descriptions by integrating multimodal feature fusion with emotionally-informed and field-acquainted knowledge.\\
\textbf{II:} We devise a  Video-Text Attribute Investigation (VTAI) module that derives multimodal concepts to generate meaningful captions. This module predicts the probability of attribute cues in each caption, encompassing objects, scenes, actions, and emotions.\\		 
\textbf{III:} We developed the Holistic Concept-Oriented Theme (HCOT) module to guide the semantic direction of captions with emotional context, minimizing misalignment. The HCOT module features two key components: the Coarse-to-Fine Block (CFB) and the Attribute Embedding Block (AEB). These components investigate query videos and their associated captions to derive fine-to-coarse concepts and ascertain each video concept's emotion and factual field.\\
\textbf{IV:} We incorporate the pre-trained model, CLIP, to enhance our proposed architecture through text and image encoding and video-to-text retrieval procedures.\\
\textbf{V:} Extensive experiments on EmVidCap \cite{30_9352546}, MSVD \cite{31_chen:acl11}, and MSR-VTT \cite{32_7780940} demonstrate that our proposed network achieves performance comparable to state-of-the-art methods in video captioning, with a specific focus on integrating emotional, factual, and holistic concepts.		
\section{RELATED WORKS}
\subsection{Video Description}
Due to their ability to generate more flexible and nuanced captions, encoder-decoder models are widely used in semantic video description. Their architecture is particularly well-suited for modeling sequential data. The encoder processes the visual data and creates a compressed representation highlighting the most important features. The decoder then interprets this compressed data to generate a sequence of words, predicting each word based on the previous ones and the overall representation. Significantly, the decoder prioritizes semantic features within the compressed representation, ensuring that the generated caption accurately reflects the video's overall meaning. One of the most important tasks for researchers is to tackle the challenge of bridging the gap between video content and its textual description. They achieve this by developing models that represent the common ground between the video's visual information and the textual content of the description. These models leverage techniques such as attention mechanisms, which enable the model to focus on specific parts of the video data most relevant to the text \cite{34_8620348}. Alternatively, they achieve this by extracting high-level semantic representations from both the video and text data, thereby capturing the core meaning of each \cite{4_9367203}, \cite{37_10.1145/3394171.3413885}. In some works, \cite{42_8917665}, \cite{44_8807239}, researchers design efficient encoder models to extract concise and informative representations from video data. These encoders specifically prioritize uncovering the semantic meaning within a single present modality in the video. In order to leverage the richness of video data, which contains multiple information channels, some research efforts  \cite{38_8447210}, \cite{39_8733019} aim to combine features from different modalities for processing by the caption generation model. Other studies, \cite{40_Zhang2020ObjectRG}, for example, investigate methods for fine-tuning pre-trained models on specific tasks to enhance performance and explore the limitations of knowledge transfer from large models across different domains.\\
In this paper, we aim to bridge the gap between video and text through high-level semantics and investigate the concept transfer capabilities of pre-trained models.
\subsection{Emotional Captioning} 	
Integrating emotion into video descriptions, \cite{30_9352546}, \cite{45_song2022contextual}, \cite{46_10418849} and considering emotional states \cite{49_10.1145/3487553.3524649}, represent emerging concepts that are still in their initial stages of development. 
 Some researchers have developed new datasets designed for video captioning that incorporate emotional dimensions \cite{30_9352546}. They also developed a system for video captioning that incorporates both factual details and emotional nuances within video content \cite{30_9352546}, \cite{48_Venugopalan_2015_ICCV}. When generating final captions, the system accurately reflects both the factual content and the emotional aspects of the video, considering their respective likelihoods.\\
In some other works, \cite{45_song2022contextual}, \cite{46_10418849}, they introduced a visual-textual attention module into the LSTM to improve the accuracy of emotional descriptions. Additionally, they integrated emotional embedding learning into the video description architecture and fine-tuned the network by combining emotional understanding with factual accuracy. A line of research, \cite{50_Achlioptas2021ArtEmisAL}, \cite{51_9428453}, introduces methods for generating captions influenced by the emotions detected in the video's visual content. This approach leverages a pre-trained emotion detection model to analyze the video and categorize the present emotions. While this approach successfully identifies emotions in videos, it categorizes them broadly (e.g., happy, sad). This limitation hinders the descriptions from capturing the full spectrum of emotions and their nuances, resulting in less detailed and impactful captions.\\ 
In this paper, we introduce the Concept Investigation Unit (CIU) to generate emotionally-informed descriptive sentences through coarse-to-fine concept exploration of the context.
			
\subsection{Vision Language Model}
Recently, numerous researchers have turned to contrastive pre-training for Vision-Language Models (VLMs). This adaptability makes them a powerful resource for addressing novel problems, \cite{58_NEURIPS2022_960a172b}, \cite{59_9878937}. Despite achieving impressive results, large VLMs require vast amounts of data and are computationally demanding. To address these challenges, researchers are exploring several key directions to enhance their efficiency. Utilizing a diverse set of prompts generated by LLMs and learning sets of tokens enhances the distinguishing features of each category.
Leveraging a diverse set of prompts generated by LLMs and utilizing learned sets of tokens that enhance the distinguishing features of each category \cite{63_10203479}, \cite{64_Parisot_2023_CVPR}, \cite{65_9879913}, developing a method to ensure that features captured in different latent spaces correspond effectively with one another \cite{61_ouali2023black}, utilizing pre-trained VLMs with minimal additional training required, even in the absence of temporal dependencies \cite{62_bain2022cliphitchhikers}, are the primary directions contributing to improvements in both accuracy and the model's ability to explain its reasoning. Vision-language models like CLIP \cite{26_radford2021learning}, ALBEF \cite{68_NEURIPS2021_50525975}, BEiT-3 \cite{69_Wang2023ImageAA}, BLIP \cite{71_unknown}, mPLUG \cite{72_li2022mplug} and some other models excelled at handling diverse tasks due to their ability to form strong connections between visual information and language. Following the development of CLIP, researchers are exploring its potential for adaptation across a broader spectrum of vision-language tasks \cite{78_10.1145/3474085.3479207}, specifically video-text retrieval \cite{77_LUO2022293}.  		
\subsection{Video-Text Retrieval}
Due to its critical role in bridging vision and language, video-text retrieval \cite{80_Zellers_2022_CVPR}, \cite{81_wang2023video}, \cite{91_Zhang_2023_ICCV}, has acquired considerable research interest as a primary task of multimodal learning, identifying an appropriate stand-in task facilitates models to learn how to associate visual and language-conveyed information. To retrieve the most relevant video based on a text prompt, prevalent video-to-text retrieval models compute a similarity matrix between visual and language information, requiring that the features extracted from both modalities correspond effectively \cite{88_Wang_2023_ICCV}, \cite{90_9710717}, this can be achieved by prioritizing the development of effective fusion mechanisms \cite{83_Yu_2018_ECCV}, introducing a sparse sampling scheme in the visual stream \cite{85_Lei_2021_CVPR}, evolving dynamic scene using a content duplicator and key-value memory as visual-language pairs  \cite{97_10.1145/3539225} and leveraging curriculum learning to adaptively train a visual encoder by utilizing temporal context \cite{92_9711165}. Following the success of CLIP \cite{26_radford2021learning}, several studies, \cite{77_LUO2022293}, \cite{87_unknown}, \cite{94_xue2023clipvip},  have employed it for video-text retrieval tasks, demonstrating state-of-the-art performance.\\
In our paper, the video-to-text retrieval module, CLIP, is utilized to obtain more accurate emotionally-informed and field-acquainted concepts by comparing video and textual features.		
\section{METHODOLOGY}
\begin{figure*}[!t]
	\centering
	\includegraphics[height = 7.0cm, width=17.5cm]{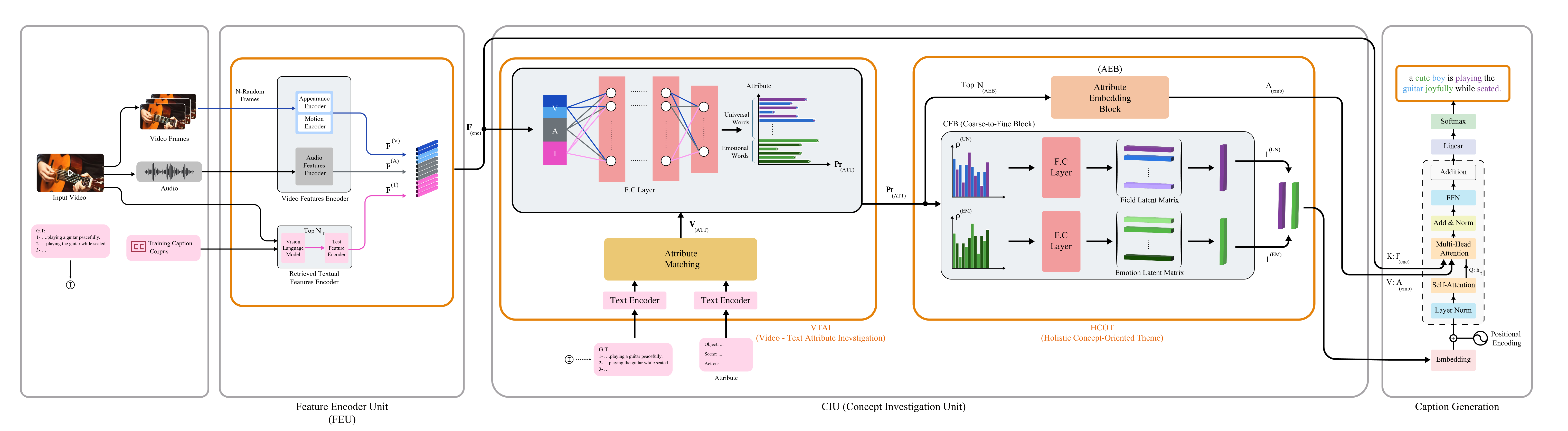}
	\caption{\footnotesize Overview of the proposed SPECTRUM. Our model utilizes the retrieved textual features with video signals to generate accurate captions. Given $\mathbf{F}_{(enc)}$ as a result of concatenation of multimodal features and  $\mathbf{A}_{(emb)}$ (\textbf{Key} matrix and \textbf{Value} matrix) extracted from the predicted probability of attribute concept $Pr^{(ATT)}$, the Pre-LN Transformer generates emotionally-informed and factually-acquainted captions for accurate description of video contents.}
	\label{fig2_structure}
\end{figure*}

\noindent Figure \ref{fig2_structure} illustrates the general architecture of our proposed model, which includes the Feature Encoders Unit (FEU), the Concept Investigation Unit (CIU), and the Knowledge Acquisition and Caption Generation component. We will provide a detailed explanation of the various components and their substructures.
\subsection{Features Encoders Unit}
Within the Feature Encoders Unit (FEU), we employ a multimodal approach that integrates visual, audio, and retrieved textual features. These multimodal features are processed within a feature representation module, where the visual and audio components are fused into a unified video feature. The combined video features and textual data are then fed into the VTAI module. The feature extraction processes implemented in the FEU will be detailed in the subsequent sub-sections.		
\subsubsection{Visual Features}
The Visual Features Encoder primarily consists of the Appearance-Based Encoder, $\Psi_{a}$, and the Motion-Based Encoder, $\Psi_{m}$.
Sampling frames from the video at regular intervals and extracting appearance features using an appropriate backbone network (e.g., Inception-ResNet-V2 \cite{133_Szegedy_Ioffe_Vanhoucke_Alemi_2017} or ViT \cite{101_dosovitskiy2020image}) yields frame-level appearance features, represented as $\mathbf{F}^{(a)} = \left\{f^a_1, f^a_2, ..., f^a_{N_a}\right\}$, where $N_{a}$ represents the number of sampled frames $\mathbf{V}_{(a)} \in \mathbb{R}^{N_{a}}$, and superscript $a$ denotes appearance. Using the appearance-based encoder, $\Psi_{a}$, the extraction of video appearance features $\mathbf{F}^{(a)} \in \mathbb{R}^{N_{a}\times d_h}$, with a model dimension of $d_h$ can be formulated as follows: 
	   \begin{equation}
	   	\label{eq_2_2}
			\mathbf{F}^{(a)} = \Psi_{a} (\mathbf{V}_{(a)}) =\Gamma ({\digamma_{a} (\mathbf{V}_{(a)}) . \mathbf{W}^{\Psi_{a}} + b^{{\Psi}_{a}}}) 
		\end{equation}			
where $\digamma_a$ is the pre-trained feature extractor model (e.g., Inception-ResNet-V2 or ViT) used for appearance extraction with the output dimension of $d_b$. The parameters $\mathbf{W}^{\Psi_{a}} \in \mathbb{R}^{d_b \times d_h}$ and ${b^{{\Psi}_{a}}} \in \mathbb{R}^{d_h}$ are learned during training, and $\Gamma$ represents a normalization layer applied within the encoder.\\
To extract motion features from the video, we apply the  Kinetics \cite{100_kay2017kinetics} pre-trained 3D ResNeXt model \cite{99_hara3dcnns}, designed to classify motion classes. The motion features are represented as: $\mathbf{F}^{(m)} = \left\{f^m_1, f^m_2, ..., f^m_{N_m}\right\}$ where $N_{m}$ is the number of snippets $\mathbf{V}_{(m)} \in \mathbb{R}^{N_m}$, and superscript $m$ denotes motion. Using motion-based encoder, $\Psi_{m}$, the extraction of video motion features $\mathbf{F}^{(m)} \in \mathbb{R}^{N_{m}\times d_h}$ with a model dimension of $d_h$, can be formulated as follows:		
		\begin{equation}\label{eq_2_3}
         \mathbf{F}^{(m)} = \Psi_{m} (\mathbf{V}_{(m)}) =\Gamma ({\digamma_{m} (\mathbf{V}_{(m)}) . \mathbf{W}^{\Psi_{m}} + b^{{\Psi}_{m}}}) 
      	\end{equation}      
where $\digamma_m$ is the pre-trained feature extractor model (e.g., 3D ResNeXt) used for motion extraction with the output dimension of $d_b$. The parameters $\mathbf{W}^{\Psi_{m}} \in \mathbb{R}^{d_b \times d_h}$ and ${b^{{\Psi}_{m}}} \in \mathbb{R}^{d_h}$ are learned during training.      
\subsubsection{Audio Features}
Inspired by the VGG architecture, the VGGish model is a pre-trained convolutional neural network (CNN) designed for audio classification tasks. It converts audio input into a semantically meaningful, high-level, compact, and computationally efficient embedding. The extracted features are represented as: 
$\mathbf{F}^{(A)} = \left\{f^A_1, f^A_2, ..., f^A_{N_A}\right\}$ where $N_A$ represents the number of semantically meaningful embedding elements learned from the audio data $\mathbf{V}_{(A)} \in \mathbb{R}^{N_A}$, and superscript $A$ denotes audio.
Using the audio encoder, the extraction of audio features $\mathbf{F}^{(A)} \in \mathbb{R}^{N_A\times d_h}$ with a model dimension of $d_h$, can be formulated as follows: 		
\begin{equation}\label{eq_2_4}
	\mathbf{F}^{(A)} = \Psi_{A} (\mathbf{V}_{(A)}) =\Gamma ({\digamma_{A} (\mathbf{V}_{(A)}) . \mathbf{W}^{\Psi_{A}} + b^{{\Psi}_{A}}}) 
\end{equation}
where in audio encoder $\Psi_{A}$, $\digamma_A$ denotes the pre-trained feature extractor model (e.g., VGGish) used for audio features extraction, with an output dimension of $d_b$. The parameters $\mathbf{W}^{\Psi_{A}} \in \mathbb{R}^{d_b \times d_h}$ and ${b^{{\Psi}_{A}}} \in \mathbb{R}^{d_h}$ are learned during training.		
\subsubsection{Retrieved Textual Features}
Using CLIP \cite{26_radford2021learning} as the video-to-text retrieval module, ground truth captions are fed into CLIP's text encoder as text queries. The embedded caption vectors are obtained as $\mathbb{X}^{(C)} = \left\{\mathbf{C}^{*}_{1}, \mathbf{C}^{*}_{2}, \cdots, \mathbf{C}^{*}_{C}\right\}$, where $\mathbf{C}^{*}_{j}$ denotes the $j^{\text{th}}$ embedded caption for $j \in [1, C]$, and $\mathbb{X}^{(C)}$ has a length of $C$, corresponding to the number of ground truth captions. For each query video processed by CLIP's image encoder, the resulting mean-pooled video feature $\mathbf{V}_{(i)}$ is compared with $\mathbb{X}^{(C)}$.		
Based on the comparison, the $N_T$ captions with the highest similarity to the query video are inputs for the textual encoder. The extracted textual features are represented as:		
\begin{equation}\label{eq_2_9}
           \mathbf{F}^{(T)} = \Psi_{T} (\mathbf{X}^{(T)}) = \left\{f^T_1, f^T_2, ..., f^T_{N_T}\right\}
\end{equation}		
where $\mathbf{F}^{(T)} \in \mathbb{R}^{N_T \times d_h}$ represents the extracted textual features, where $\Psi_{T}$ is the textual encoder. The set $\mathbf{X}^{(T)}$ contains the top $N_T$ individual captions with the highest semantic similarity to the query video and is defined as follows:				
\begin{equation}\label{eq_2_1}
	\mathbf{X}^{(T)} = Compare (\mathbf{V}_{(i)}, \mathbb{X}^{(C)}) = \Phi (Rnk({Sim}(\mathbf{V}_{(i)}, \mathbb{X}^{(C)})))
\end{equation}
where $\mathbf{V}_{(i)}$ is the $i^{th}$ query video. The \textit{Sim} function computes the similarity scores between the query video and the captions, resulting in a similarity vector of dimension $\lvert \mathbb{C} \rvert$. The \textit{Rnk} function ranks these scores in order, and $\Phi(\cdot)$ selects the $N_T$ retrieved semantic vectors with the highest similarity scores. Finally, textual feature extraction is achieved using GloVe embeddings.		       
\subsection{Concept Investigation Unit}
The Concept Investigation Unit (CIU) is structured around two critical phases: the video-text attribute investigation stage and the holistic concept orientation stage. These phases are crafted to produce semantically rich and meaningful captions. In the video-text attribute investigation stage, the model conducts a multimodal semantic analysis to assess the likelihood of emotional and factual words in the ground truth captions. The training process within this stage enhances the model's functionality by refining its ability to accurately identify concepts based on the probabilities associated with emotional and factual vocabularies. On the other hand, the holistic concept orientation stage focuses on specifying the conceptual orientation of captions. It aims to generate emotionally-informed and field-acquainted sentences, thereby minimizing the emergence of semantically unrelated descriptions. This stage further refines the model's ability to align attribute concepts with descriptive captions.                 
\subsubsection{Visual-Text Attribute Investigation}
In this stage, we introduce the Visual-Text Attribute Investigation (VTAI) module, instrumental in deriving multimodal concepts crucial in generating meaningful captions. The concatenated representation of visual, audio, and textual features, denoted as  $\mathbf{F}_{(enc)} = [\mathbf{F}^{(V)}; \mathbf{F}^{(A)}; \mathbf{F}^{(T)}]$  is fed into VTAI module. Within this framework, the VTAI predicts the probability of attribute cues, $Pr^{(ATT)}$, for each video, enhancing the model’s ability to generate contextually relevant and semantically rich descriptions.
Both emotional and factual words constitute the attribute cues within our model. It is important to note that the set of factual words may vary each time, influenced by changes in their repetition frequency and the predetermined threshold. This dynamic selection ensures that the model remains adaptable and responsive to varying textual contexts, thereby enhancing its capacity to generate contextually appropriate captions.\\
The predicted probability of the $i^{th}$ attribute concept is represented as:       
\begin{equation}\label{eq_2_12}
       Pr^{(ATT)} = \Lambda (x \mathbf{W}^{(ATT)} + b^{(ATT)})
\end{equation}
where $p_{{r}_{i}} \in Pr^{(ATT)}$ denotes the predicted probability for the specific attribute. Here, $x \in {\mathbb{R}^{3d_h}}$ signifies the input to the module comprising concatenated visual, audio, and textual features. The function $\Lambda$ represents the activation function used in the neural network. Moreover, $\mathbf{W}^{(ATT)}$ and $b^{(ATT)}$  are the trainable tensors, specifically the weights and biases, respectively, which are optimized during the training process.\\
The VTAI module is trained using a normalized binary cross-entropy loss, introducing an associated MultiModal Semantic (M2S) extraction loss defined as follows:           
  \begin{equation}\label{eq_9}
    \mathcal{L}_{(M2S)} = -\frac{1}{N_{(ATT)}} \mathbf{\sum_{i=1}^{N_{(ATT)} }} (v_i\log ({p_{r}}_i) + (1-v_i)\log (1-{p_{r}}_i))
  \end{equation}
where $v_i$  denotes the ground truth label for the $i^{th}$ attribute concept, and $p_{{r}_{i}}$ is the predicted probability of the $i^{th}$ attribute concept as produced by the VTAI module. This formulation aims to minimize the discrepancy between the predicted probabilities and actual labels across all attribute concepts, thereby enhancing the module's ability to extract multimodal semantic attributes accurately.                  
\subsubsection{Holistic Concept-Oriented Theme}  
The Holistic Concept-Oriented Theme (HCOT) module determines the orientation of descriptive sentences, incorporating emotionally-informed and field-acquainted references. Pre-awareness of the orientation related to a video's emotional and contextual nuances prioritizes relevant semantic words, ensuring alignment with the content description sentences. The HCOT module comprises two critical blocks: the Coarse-to-Fine Block (CFB) and the Attribute Embedding Block (AEB). These blocks are integral to refining the thematic accuracy and enhancing the semantic coherence of the generated captions, thereby facilitating a deeper connection between the multimodal inputs and the expressed descriptions.       
\paragraph*{Coarse-to-Fine Block (CFB)}:
In this phase, the model examines query videos and their associated captions, employing a fine-to-coarse conceptual analysis to determine each video's predominant emotions and overarching themes.
Investigating all captions associated with each individual video, the model identifies the presence of fine-grained emotional vocabularies. The distribution of these words is then analyzed to assign a coarse emotional concept to the video. The repetition frequency of fine-grained emotional words classifies the video into appropriate categories. On the other hand, according to the different fields assigned to the MSR-VTT dataset \cite{32_7780940}, all video queries can be grouped into 20 distinct fields, each specifying the discipline of factual words. This dual approach enables a comprehensive understanding of the videos' emotional and thematic content.
Leveraging the assigned field and emotional characteristics of each individual video and utilizing associated coarse-to-fine trainable matrices, $\mathbf{W}^{(FLD)}$,  $\mathbf{W}^{(CTG)}$, the generated captions for each video accurately reflect the content, where "FLD" and "CTG" denote fields and categories, respectively. 
Given the close relationship between video concepts and their associated thematic content, to more accurately extract holistic attribute-oriented concepts, the trainable matrices for factual and emotional concepts, $\mathbf{W}^{(FCT)}$,  $\mathbf{W}^{(EM)}$ are modulated by their corresponding probabilities,  $\rho^{(FCT)}$ and $\rho^{(EM)}$. Consequently, the latent holistic attribute vectors are calculated as follows:         
         \begin{subequations}\label{eq_9_3}
         	\begin{align}
         	l_H^ {(FCT)} = {\rho^{(FCT)}}.{\mathbf{W}^{(FLD)}}.{\mathbf{W}^{(FCT)}}\label{eq_9_3A}\\
         	l_H^ {(EM)} = {\rho^{(EM)}}.{\mathbf{W}^{(CTG)}}.{\mathbf{W}^{(EM)}}\label{eq_9_3B}
         \end{align}
         \end{subequations}              
where $l_H^ {(FCT)} \in \mathbb{R}^{d_h}$, and $l_H^ {(EM)} \in \mathbb{R}^{d_h}$  represent the latent holistic vectors for factual and emotional attributes, respectively;               
\paragraph*{Attribute Embedding Block}:
Based on  $Pr^{(ATT)}$, the features of the top emotionally-informed and field-acquainted concepts with the highest probabilities $N_{(ATT)}$ are extracted and semantically embedded.\\  
\subsection{Knowledge Acquisition and Caption Generation}
In caption generation, each word $c^*_t \in \mathbf{C}^{*}_{j}$ in the descriptive sentence depends on the video features, $[\mathbf{F}^{(V)}; \mathbf{F}^{(A)}; \mathbf{F}^{(T)}]$, and prior predicted words $c_{0:t-1}$. After feeding the input embeddings and video features into the caption decoder, which consists of $N_{(L)}$ Transformer decoder blocks, the next word $c_t$ is predicted.      
       \begin{equation}\label{eq_2_22}
       	c_t \sim p{\left(c_t|c_{0:t-1}, [\mathbf{F}^{(V)}; \mathbf{F}^{(A)}; \mathbf{F}^{(T)}]\right)}
       \end{equation}       
where $p{\left(c_t|c_{0:t-1}, [\mathbf{F}^{(V)}; \mathbf{F}^{(A)}; \mathbf{F}^{(T)}]\right)} \in \mathbb{R} ^ {\lvert \mathcal{V} \rvert}$ represents the predicted probability distribution of the current word over the entire vocabulary $\mathcal{V}$
       .\\
Given the ground-truth caption $\mathbf{C}^{*}_{j}$, the cross-entropy loss is employed to train the entire 
model and optimize the caption generation process.       
       \begin{equation}\label{eq_2_32}
       	\mathcal{L}_{(CAP)} = - \sum_{t=1}^{T}  \log \left(p{\left(c^*_t|c^*_{0:t-1}, [\mathbf{F}^{(V)}; \mathbf{F}^{(A)}; \mathbf{F}^{(T)}]\right)}\right)
       \end{equation}   
Clearly, word and positional embeddings are two essential components at each time step $t^{th}$, which significantly enhance the quality of the captions generated by the model. Moreover, in our approach, the latent holistic attribute vectors $l_H^ {(FCT)}$ and $l_H^ {(EM)}$ are aggregated with the input embeddings $I_{(emb)} \in \mathbb{R}^{d_h}$.       	
       	\begin{equation}\label{eq_9_14}
       		I_{(emb)} = Emb(c_{0:t-1},  l_H^ {(FCT)},  l_H^ {(EM)})
       	\end{equation}        	 
To facilitate hyperparameter tuning, we adopt a variant of the Transformer architecture, specifically the Transformer with Pre-Layer Normalization (Pre-LN Transformer) introduced in \cite{103_xiong20b}, as the backbone for the caption decoder. The decoder is fed with the embedding module $I_{(emb)}$, video and textual features $\mathbf{F}_{(enc)}$, and the embedded attribute vector $\mathbf{A}_{(emb)}$.\\
The cross-modal interactions are modeled using scaled dot-product attention, formulated as follows:      	
       	 \begin{equation}\label{eq_13}
       		{Attention}_{(Pre-LN)_{i}}(\mathbf{Q}_i, \mathbf{K}_i, \mathbf{V}_i) =
       		\mathrm{Softmax}\left( \frac{\mathbf{Q}_i \mathbf{K}^T_i}{\sqrt{d_k}} \right)\mathbf{V_i}
       	\end{equation}       
In this context, the concatenation of  $\mathbf{F}_{(enc)}$, $\mathbf{A}_{(emb)}$ are considered as the key (K) and value (V) matrices, while the query (Q) is derived from the outputs of the self-attention layer, serving as a hidden state, $\mathbf{h}_{t} \in \mathbb{R}^{d_h}$, representation.\\
Integrating feature encoders and concept investigation units with the Pre-LN Transformer constitutes our proposed model. This model is designed to optimize the caption generation loss $\mathcal{L}_{(CAP)}$ and multimodal semantic extraction loss $\mathcal{L}_{(M2S)}$ for precise knowledge acquisition. The overall loss function for model tuning is formulated as follows:   
       \begin{equation}\label{eq_14}
      	\mathcal{L} = \mathcal{L}_{(M2S)} + \mathcal{L}_{(CAP)}
       \end{equation}

\section{EXPERIMENT}     
    \subsection{Datasets}
    We evaluate our model on three benchmark datasets: EmVidCap \cite{30_9352546}, MSVD \cite{31_chen:acl11}, and MSR-VTT \cite{32_7780940}. The full \textbf{EmVidCap} dataset (comprising  {EmVidCap-S} with 10-second videos and an average of 7 tokens per caption, and {EmVidCap-L} with 23-second videos and an average of 11 tokens per caption) contains 27,567 captions and 1381 videos for training, and 11,138 captions over 516 videos for testing. Our evaluation is conducted on the complete EmVidCap dataset. The \textbf{MSVD} dataset consists of 1,970 videos annotated with approximately 40 captions each. We follow the split provided in \cite{48_Venugopalan_2015_ICCV} and \cite{110_yao2015describing}, using 1,200 videos for training, 100 for validation, and 670 for testing. The \textbf{MSR-VTT} dataset contains 10,000 videos, each paired with 20 captions and one of 20 field tags. We adhere to the official split \cite{32_7780940}, utilizing 6,513 video clips for training, 497 for validation, and 2,990 for testing.    
\subsection{Evaluation Metrics}
To assess the performance of the model, we employ various evaluation techniques. For video captioning, we utilize four widely recognized factual metrics: \textbf{BLUE} \cite{111_10.3115/1073083.1073135}, \textbf{METEOR} \cite{112_lavie-agarwal-2007-meteor}, \textbf{ROUGE-L} \cite{113_lin2004rouge}, and \textbf{CIDEr} \cite{114_7299087} to measure the accuracy and coherence of the generated sentences. Additionally, following \cite{30_9352546},we adopt two emotion metrics, $\mathbf{{Acc_{sw}}}$ and $\mathbf{{Acc_{c}}}$, to effectively evaluate the accuracy of emotion at the word level and sentence level, respectively. Furthermore, novel holistic metrics, \textbf{BFS} and \textbf{CFS}, introduced in \cite{30_9352546} combine BLEU and CIDEr with emotion metrics for a more comprehensive evaluation.
 \begin{table*}[!t]
	\centering
	\caption{Primary Comparison of Multimodal Features on the EmVidCap Dataset with Corresponding Results}
	\label{table_Comparison on Emotional Video Captioning}
	\begin{adjustbox}{max width=\textwidth}
		\begin{tabular}{|>{\centering\arraybackslash}m{2.5cm}|>{\centering\arraybackslash}m{2.0cm}|>{\centering\arraybackslash}m{1.5cm}|>{\centering\arraybackslash}m{1.8cm}|>{\centering\arraybackslash}m{1.6cm}|>{\centering\arraybackslash}m{1.6cm}|>{\centering\arraybackslash}m{1.6cm}|>{\centering\arraybackslash}m{1.6cm}|>{\centering\arraybackslash}m{2.0cm}|>{\centering\arraybackslash}m{2.2cm}|>{\centering\arraybackslash}m{1.6cm}|>{\centering\arraybackslash}m{1.2cm}|>{\centering\arraybackslash}m{1.2cm}|>{\centering\arraybackslash}m{1.2cm}|>{\centering\arraybackslash}m{1.2cm}|>{\centering\arraybackslash}m{1.2cm}|}
			\hline
			
			\multirow{3}{*}{{Model}} & \multicolumn{3}{c|}{\multirow{2}{*}{{Feature}}} & \multicolumn{8}{c|}{\multirow{2}{*}{{Factual Metrics}}} & \multicolumn{2}{c|}{\parbox[c][2.5em][c]{1.8cm}{\centering \vspace{1.4em}{Emotional \\ Metrics}}} & \multicolumn{2}{c|}{\parbox[c][2.5em][c]{1.8cm}{\centering \vspace{1.4em}{Holistic \\ Metrics}}} \\ 
			& \multicolumn{3}{c|}{} & \multicolumn{8}{c|}{} & \multicolumn{2}{c|}{} & \multicolumn{2}{c|}{} \\ \cline{2-16}
			& {Visual} & {Audio} & {Retrieval} & {BLUE-1} & {BLUE-2} & {BLUE-3} & {BLUE-4} & {METEOR} & {ROUGE\_L} & {CIDEr} & {Sum} & {Acc\textsubscript{sw}} & {Acc\textsubscript{c}} & {BFS} & {CFS} \\ \hline

			FT \cite{30_9352546}& R152 & $-$ & $-$ & 67.6 & 47.2 & 32 & 21.6 & 20.4 & 43.1 & 29 & 114.1 & 51.2 & 49.6 & 37.6 & 33.3 \\ \hline
			CANet \cite{45_song2022contextual} & R101 + 3D-RN & $-$ & $-$ & 68.1 & 47.7 & 32.9 & 22.5 & 19.7 & 43.7 & 34.5 & 120.4 & 53.7 & 52.7 & 38.8 & 38.2 \\ \hline
			SA \cite{116_8578893} & R101 + 3D-RN & $-$ & $-$ & 68.4 & 48.3 & 33.3 & 22.4 & 19.8 & 44.1 & 32.4 & 118.7 & 48.4 & 45.5 & 37.8 & 35.3 \\ \hline
			SGN \cite{126_ryu2021semantic} & R101 + 3D-RN & $-$ & $-$ & 68.7 & 48.9 & 34.2 & 24 & 20.1 & 44.8 & 35.5 & 124.4 & 50.4 & 48.6 & 39.1 & 38.3 \\ \hline
			VEIN \cite{46_10418849} & CLIP & $-$ & $-$ & 72.1 & 52.8 & 37.9 & 27.1 & 21.6 & 46.8 & 39.4 & 134.9 & 59 & 57.6 & 43.6 & 43.1 \\ \hline
			\multirow{6}{*}{{SPECTRUM}} & R101 + 3D-RN & VGGish & CLIP - ViT-B/32 & 77.4 & 58.1 & 42.2 & 29.5 & 22.9 & 49.1 & 42.8 & 144.3 & 79.6 & 78.9 & 50.9 & 50.1 \\ \cline{2-16}
			& IRv2 + 3D-RN & VGGish & CLIP - ViT-B/32 & 77.4 & 58.4 & 43.0 & 30.8 & 23.4 & 49.7 & 45.0 & 148.9 & \textbf{83.0} & \textbf{82.4} & 52.3 & 52.5 \\ \cline{2-16}
			& RN50*4 + 3D-RN & VGGish & CLIP - ViT-B/32 & 78.5 & 60.2 & 44.2 & 31.6 & 23.3 & 50.1 & 46.5 & 151.5 & 79.9 & 79.3 & 52.5 & 53.1 \\ \cline{2-16}
			& ViT-B/32 + 3D-RN & VGGish & CLIP - ViT-B/32 & \textbf{80.0} & \textbf{61.7} & \textbf{46.1} & \textbf{32.9} & \textbf{24.1} & \textbf{50.9} & \textbf{47.8} & \textbf{155.8} & 79.6 & 78.7 & \textbf{53.7} & \textbf{54.1} \\ \hline
		\end{tabular}
	\end{adjustbox}
\end{table*}  
\subsection{Implementation Details}
In our proposed model, within the Features Encoders Unit (FEU), the image encoder from CLIP's ViT-B/32 and Inception-ResNet-v2 (IRv2) are employed to encode appearance visual features, while Kinetics pre-trained 3D ResNeXt is utilized to extract motion visual features. For audio features extraction, we employ the AudioSet pre-trained VGGish model. Video-to-text retrieval is performed using CLIP\textsubscript{ViT-B/32}, and textual feature extraction is achieved using GloVe embeddings.\\
The model is trained over 50 epochs with an initial learning rate of $lr=5e-7$. Tokenized, lowercased, and truncated sentences with a maximum length of $L_{(max)} = 30$ are used. The training is performed with video-caption pairs in batches of 128. Experiments are conducted using PyTorch on an NVIDIA GeForce RTX 2080 Ti GPU, with the model optimized using the ADAM optimizer \cite{115_article} and an L2 weight decay of 0.001. The learning rate is decayed by a factor of 0.9 after each epoch. During the testing phase, beam search with a beam size of 5 is employed for inference.             
 \subsection{Comparison with State-of-the-art Methods}
 \subsubsection{Benchmarking Methods for Emotional Video Captioning} 
In this section, the performance of the SPECTRUM model is evaluated by comparing it with established methods in Emotional Video Captioning. Using the EmVidCap dataset, the model's effectiveness is benchmarked against the latest advanced approaches to determine its performance relative to existing standards. The results, presented in Table \ref{table_Comparison on Emotional Video Captioning}, show that SPECTRUM, which integrates multiple visual feature encoders, significantly outperforms previous state-of-the-art models across emotional, factual, and holistic evaluation metrics. 
Table \ref{table_Comparison on Emotional Video Captioning} demonstrates the superior performance of our model. For instance, when compared to the VEIN model, SPECTRUM enhances the scores across multiple evaluation metrics:  Sum improves from 134.9 to 155.8, BFS from 43.6 to 53.7, and CFS from 43.1 to 54.1. This represents significant percentage improvements of $\%15.49$, $ \%23.17$, and $\%25.52$ in Sum, BFS, and CFS, respectively. Needless to say, the choice of visual feature encoder significantly contributes to the performance enhancements observed.\\
 Employing the same visual features as the SGN model, our multimodal architecture achieves substantial improvements on the EmVidCap dataset, with gains of $\%16$, $\%30.18$, and $\%30.81$ in the Sum, BFS, and CFS metrics, respectively. This remarkable progress highlights the superiority of our multimodal architecture.  
 \subsubsection{Benchmarking Methods for Factual Video Captioning} 
To assess the performance of the SPECTRUM on different datasets, it was tested on well-known semantic benchmarks:  MSVD \cite{31_chen:acl11} and MSR-VTT \cite{32_7780940}, after removing emotional learning components for an unbiased comparison with current state-of-the-art (SoTA) methods. The results, shown in Table \ref{table_Comparison on Factual Video Captioning}, indicate that SPECTRUM consistently surpasses all other models in all metrics across both datasets.\\
This emphasizes the effectiveness of incorporating multimodal information for accurate concept recognition and caption generation. These findings affirm SPECTRUM's ability to adeptly handle both emotional and factual aspects of video captioning tasks.
\begin{table}[!t]
	\centering
	\caption{Performance Comparison of Factual Video Captioning on MSVD and MSR-VTT Datasets. "V":Visual Modality. "A":Audio Modality. "T":Textual Modality.}
	\label{table_Comparison on Factual Video Captioning}
	\begin{adjustbox}{max width=\textwidth}
		\scriptsize
		\begin{tabular}{|>{\centering\arraybackslash}m{1.6cm}|>{\centering\arraybackslash}m{0.6cm}|>{\centering\arraybackslash}m{0.6cm}|>{\centering\arraybackslash}m{0.6cm}|>{\centering\arraybackslash}m{0.6cm}|>{\centering\arraybackslash}m{0.7cm}|>{\centering\arraybackslash}m{0.7cm}|>{\centering\arraybackslash}m{0.6cm}|>{\centering\arraybackslash}m{0.6cm}|>{\centering\arraybackslash}m{0.6cm}|>{\centering\arraybackslash}m{0.6cm}|>{\centering\arraybackslash}m{0.7cm}|}
			\hline
			\multirow{3}{*}{{Model}} & \multirow{3}{*}{{Year}} & \multicolumn{5}{c|}{\multirow{2}{*}{{MSVD}}} & \multicolumn{5}{c|}{\multirow{2}{*}{{MSR-VTT}}} \\ 
			&  & \multicolumn{5}{c|}{} & \multicolumn{5}{c|}{} \\ \cline{3-12}
			&  & {B-4} & {M} & {R} & {C} & {Sum} & {B-4} & {M} & {R} & {C} & {Sum} \\ \hline
			
			STG-KD \cite{123_Pan_2020_CVPR} & 2020 & 52.2 & 36.9 & 73.9 & 93.0 & 256 & 40.5 & 28.3 & 60.9 & 47.1 & 176.8 \\ \hline
			ORG-TRL \cite{124_9156538} & 2020 & 54.3 & 36.4 & 73.9 & 95.2 & 259.8 & 43.6 & 28.8 & 62.1 & 50.9 & 185.4 \\ \hline
			SAAT \cite{125_zheng2020syntax}& 2020 & 46.5 & 33.5 & 69.4 & 81.0 & 230.4 & 40.5 & 28.2 & 60.9 & 49.1 & 178.7 \\ \hline
			SGN \cite{126_ryu2021semantic} & 2021 & 52.8 & 35.5 & 72.9 & 94.3 & 255.5 & 40.8 & 28.3 & 60.8 & 49.5 & 179.4 \\ \hline
			MGCMP \cite{127_chen2021motion} & 2021 & 55.8 & 36.9 & 74.5 & 98.5 & 265.7 & 41.7 & 28.9 & 62.1 & 51.4 & 184.1 \\ \hline
			HRNAT \cite{128_gao2021hierarchical} & 2022 & 55.7 & 36.8 & 74.1 & 98.1 & 264.7 & 42.1 & 28.0 & 61.6 & 48.2 & 179.9 \\ \hline
			SHAN \cite{4_9367203} & 2022 & 54.3 & 35.3 & 72.2 & 91.3 & 253.1 & 39.7 & 28.3 & 60.4 & 49.0 & 177.4 \\ \hline
			PDA \cite{129_wang2021pos} & 2022 & 58.7 & 37.6 & 74.8 & 100.3 & 271.4 & 43.8 & 28.8 & 62.1 & 51.2 & 185.9 \\ \hline
			LSRT \cite{1_9741388} & 2022 & 55.6 & 37.1 & 73.5 & 98.5 & 264.7 & 42.6 & 28.3 & 61.0 & 49.5 & 181.4 \\ \hline
			TVRD \cite{3_9762283} & 2022 & 50.5 & 34.5 & 71.7 & 84.3 & 241.0 & 43.0 & 28.7 & 62.2 & 51.8 & 185.7 \\ \hline
			RSFD \cite{130_zhong2023refined} & 2023 & 51.2 & 35.7 & 72.9 & 96.7 & 256.5 & 43.8 & 29.3 & 62.3 & 53.1 & 188.5 \\ \hline
			VEIN \cite{46_10418849} & 2023 & 55.7 & 37.6 & 74.4 & 98.9 & 266.6 & 44.1 & 30.0 & 62.9 & 55.3 & 192.3 \\ \hline
			{{SPECTRUM} (V)}  & $-$ & 55.5 & 38.2 & 74.4 & 101.5 & 269.6 & 46.7 & 30.7 & 63.9 & 58.3 & 199.6 \\ \hline
			{{SPECTRUM} (V + A)}  & $-$ & 55.9 & 38.4 & 74.8 & 102.1 & 271.2 & 46.9 & 30.7 & 64 & 58.5 & 200.1 \\ \hline
			{{SPECTRUM} (V + T)}  & $-$ & 57.6 & 38.6 & 75.0 & 102.6 & 273.8 & 47.4 & 30.9 & 64.2 & 58.8 & 201.3 \\ \hline
			{{SPECTRUM} (V + A + T)} & $-$ & \textbf{59.3} & \textbf{38.8} & \textbf{75.2} & \textbf{104.3} & \textbf{277.6} & \textbf{48.1} & \textbf{31.4} & \textbf{64.6} & \textbf{59.1} & \textbf{203.2} \\ \hline
		\end{tabular}
	\end{adjustbox}
\end{table}
 \subsection{Ablation Study} 
We conducted a series of ablation studies on the EmVidCap dataset to thoroughly evaluate the impact of various components within our proposed method.
\subsubsection{Impact of Various Image Feature Extractors}
The Inception-ResNet-V2 (IRv2) and CLIP models were compared in our video captioning experiments to assess their effectiveness in encoding image content. While IRv2 has been a traditional choice, the CLIP model has shown stronger capabilities in robust image representation. This impact is evident in Table \ref{table_Image-Backbone_HCOT_Modality}, where the choice of image feature extractor significantly affects the model’s ability to interpret both emotional and factual video content. Experimental results indicate that using CLIP leads to notable improvements across emotional, factual, and holistic metrics. The baseline model referenced in Table \ref{table_Image-Backbone_HCOT_Modality} does not incorporate any of the designed blocks from our proposed architecture (i.e., VTAI and HCOT). Instead, it solely relies on a transformer to decode and translate video features into textual representations.
\begin{table*}[!t]
	\centering
	\caption{Impact of Various Components of SPECTRUM on the EmVidCap Dataset. "IRV2":Inception-ResNet-V2. "CFB":Coarse to Fine Block. "AEB":Attribute Embedding Block.}
	\label{table_Image-Backbone_HCOT_Modality}
	\begin{adjustbox}{max width=\textwidth}
		\begin{tabular}{|>{\centering\arraybackslash}m{1cm}|>{\centering\arraybackslash}m{2.7cm}|>{\centering\arraybackslash}m{2.4cm}|>{\centering\arraybackslash}m{1.3cm}|>{\centering\arraybackslash}m{1.3cm}|>{\centering\arraybackslash}m{1.4cm}|>{\centering\arraybackslash}m{2.5cm}|>{\centering\arraybackslash}m{1.6cm}|>{\centering\arraybackslash}m{2cm}|>{\centering\arraybackslash}m{2cm}|>{\centering\arraybackslash}m{1.3cm}|>{\centering\arraybackslash}m{1.3cm}|>{\centering\arraybackslash}m{1.2cm}|>{\centering\arraybackslash}m{1.1cm}|>{\centering\arraybackslash}m{1.3cm}|>{\centering\arraybackslash}m{1.3cm}|}			
			\hline

			\multirow{3}{*}{\centering \makecell{{Exp.} \\ {No.}}} & \multirow{3}{*}{\centering \makecell{{Image} \\ {Feature} \\ {Extractor}}} & \multirow{3}{*}{\centering \makecell{{HCOT} \\ {Module}}} &
			\multicolumn{3}{c|}{\multirow{2}{*}{{Modality}}} &
			\multirow{3}{*}{\centering {Model}} &
			\multicolumn{5}{c|}{\multirow{2}{*}{{Factual Metrics}}} & \multicolumn{2}{c|}{\parbox[c][2.5em][c]{2.5cm}{\centering \vspace{1.4em}{Emotional \\ Metrics}}} & \multicolumn{2}{c|}{\parbox[c][2.5em][c]{2.5cm}{\centering \vspace{1.4em}{Holistic \\ Metrics}}} \\ 
			&  &  & \multicolumn{3}{c|}{} & & \multicolumn{5}{c|}{} & \multicolumn{2}{c|}{} & \multicolumn{2}{c|}{} \\ \cline{4-6} \cline{8-16}
			&  &  & {Visual} & {Audio} & {Textual} & & {BLEU-4} & {METEOR} & {ROUGE-L} & {CIDEr} & {SUM} & {Acc\textsubscript{sw}} & {Acc\textsubscript{c}} & {BFS} & {CFS} \\ \hline

			1  & \multirow{8}{*}{IRv2} & CFB + AEB & $\checkmark$ & $\checkmark$ & $\checkmark$ & $-$ & \textbf{30.8} & \textbf{23.4} & \textbf{49.7} & \textbf{45.0} & \textbf{148.9} & \textbf{83.0} & \textbf{82.4} & \textbf{52.3} & \textbf{52.5} \\ \cline{1-1} \cline{3-16}
			2  &  & CFB & $\checkmark$ & $\checkmark$ & $\checkmark$ & $-$ & 27.6 & 22.3 & 48.0 & 41.9 & 139.8 & 67.3 & 66.5 & 48.9 & 46.9 \\ \cline{1-1} \cline{3-16}
			3  &  & AEB & $\checkmark$ & $\checkmark$ & $\checkmark$ & $-$ & 26.9 & 21.0 & 46.6 & 40.8 & 135.3 & 65.3 & 64.5 & 47.9 & 45.6 \\ \cline{1-1} \cline{3-16}
			4 &  & {CFB + AEB} & $\checkmark$ & $\checkmark$ & $\times$ & $-$ &  25.9 & 20.8 & 45.5 & 39.1 & 131.3 & 61.8 & 60.9 & 46.5 & 43.6 \\ \cline{1-1} \cline{3-16}
			5  &  & {CFB + AEB}  & $\checkmark$ & $\times$ & $\checkmark$ & $-$ & 26.9 & 21.6 & 46.9 & 39.7 & 135.1 & 63.4 & 62.6 & 47.7 & 44.4 \\ \cline{1-1} \cline{3-16}
			6 &  &  {CFB + AEB} & $\checkmark$ & $\times$ & $\times$ & $-$ & 25.5 & 19.6 & 44.8 & 37.3 & 127.2 & 57.9 & 56.6 & 45.2 & 41.3 \\ \cline{1-1} \cline{3-16}
			7 &   & $-$ & $-$ & $-$ & $-$ & Baseline & 24.6 & 19.3 & 44.6 & 37.1 & 125.6 & 57.2 & 55.8 & 44.7 & 41.0 \\ \hline
	
			8  & \multirow{8}{*}{CLIP-ViT-B/32} & CFB + AEB & $\checkmark$ & $\checkmark$ & $\checkmark$ & {SPECTRUM} & \textbf{32.9} & \textbf{24.1} & \textbf{50.9} & \textbf{47.8} & \textbf{155.8} & \textbf{79.6} & \textbf{78.7} & \textbf{53.7} & \textbf{54.1} \\ \cline{1-1} \cline{3-16}
			9  &  & CFB & $\checkmark$ & $\checkmark$ & $\checkmark$ & $-$ & 30.8 & 23.0 & 48.9 & 45.7 & 148.4 & 73.8 & 71.5 & 49.4 & 51.1 \\ \cline{1-1} \cline{3-16}
			10  &  & AEB & $\checkmark$ & $\checkmark$ & $\checkmark$& $-$ & 29.4 & 22.3 & 48.1 & 43.1 & 143.0 & 64.9 & 64.1 & 47.5 & 47.4 \\ \cline{1-1} \cline{3-16}
			
			11 &  & {CFB + AEB} & $\checkmark$ & $\checkmark$ & $\times$ & $-$ & 30.6 & 22.7 & 48.9 & 43.7 & 145.9 & 66.4 & 63.9 & 48.6 & 48.0 \\ \cline{1-1} \cline{3-16}
			12  &  & CFB + AEB & $\checkmark$ & $\times$ & $\checkmark$ & $-$ & 31.4 & 23.3 & 49.7 & 44.8 & 149.2 & 68.8 & 66.2 & 49.6 & 49.3 \\ \cline{1-1} \cline{3-16}
			13  &  & CFB + AEB & $\checkmark$ & $\times$ & $\times$ & $-$ & 28.8 & 22.5 & 48.3 & 42.8 & 142.4 & 63.1 & 61.9 & 46.9 & 46.7 \\ \cline{1-1} \cline{3-16}
			14  & & $-$ & $-$ & $-$ & $-$ & Baseline & 26.6 & 20.7 & 46.5 & 40.5 & 134.3 & 59.3 & 57.7 & 44.5 & 44.1 \\ \hline
		\end{tabular}
	\end{adjustbox}
\end{table*}
\subsubsection{Impact of Emotion and Field Embeddings within the CFB}
Consistent with previous discussions, each sub-component of the HCOT module (i.e., CFB and AEB) independently enhances model performance. As shown in Table \ref{table_Image-Backbone_HCOT_Modality}, eliminating the CFB led to a decline in emotional and factual metrics. The analysis of Table \ref{table_embedding_emotional_field_CFB} further reveals that emotion and field embeddings play distinct yet complementary roles in maintaining the performance of the CFB. The exclusion of the field embedding (Exp. 2) results in only minor performance degradation. In contrast, removing emotion embedding (Exp. 3) leads to substantial performance losses, particularly in emotional and holistic metrics, indicating that emotional information is essential for generating contextually relevant outputs and preserving overall model coherence. Furthermore, the simultaneous removal of both embeddings (Exp. 4) results in a marked performance drop across all metrics. These results highlight the importance of embedding strategies in enhancing multimodal architectures. 
\begin{table}[!t]
	\centering
	\caption{Effect of Emotion and Field Embedding Procedures in CFB.}
	\label{table_embedding_emotional_field_CFB}
	\begin{adjustbox}{max width=\textwidth}
		\scriptsize
		\begin{tabular}{|>{\centering\arraybackslash}m{0.6cm}|>{\centering\arraybackslash}m{2.1cm}|>{\centering\arraybackslash}m{0.6cm}|>{\centering\arraybackslash}m{0.6cm}|>{\centering\arraybackslash}m{0.6cm}|>{\centering\arraybackslash}m{0.6cm}|>{\centering\arraybackslash}m{0.8cm}|>{\centering\arraybackslash}m{0.8cm}|>{\centering\arraybackslash}m{0.65cm}|>{\centering\arraybackslash}m{0.65cm}|>{\centering\arraybackslash}m{0.65cm}|}
			\hline

			\multirow{3}{*}{\centering \makecell{{Exp.} \\ {No.}}} & \multirow{3}{*}{\centering \makecell{{CFB} \\ {Module}}} &  \multicolumn{5}{c|}{\multirow{2}{*}{{Factual Metrics}}} & 
			\multicolumn{2}{c|}{\parbox[c][1.2em][c]{1.2cm}{\centering \vspace{1.4em}{Emotional \\ Metrics}}} & 
			\multicolumn{2}{c|}{\parbox[c][1.2em][c]{1.2cm}{\centering \vspace{1.4em}{Holistic \\ Metrics}}} \\ 
			&  & \multicolumn{5}{c|}{} & \multicolumn{2}{c|}{} & \multicolumn{2}{c|}{}  \\ \cline{3-11}
			&  & {B-4} & {M} & {R} & {C} & {SUM} & {Acc\textsubscript{sw}} & {Acc\textsubscript{c}} & {BFS} & {CFS} \\ \hline

			1 & {{SPECTRUM}} & \textbf{32.9} & \textbf{24.1} & \textbf{50.9} & \textbf{47.8} & \textbf{155.8} & \textbf{79.6} & \textbf{78.7} & \textbf{53.7} & \textbf{54.1} \\ \hline
			2 & w/o Field Embedding & 32.2 & 23.5 & 49.8 & 47.4 & 152.9 & 78.2 & 77.5 & 52.8 & 53.5 \\ \hline
			3 & w/o Emotion Embedding & 30.1 & 22.8 & 48.7 & 44.8 & 146.3 & 69.6 & 68.6 & 48.5 & 49.6 \\ \hline
			4 & w/o Field \& Emotion Embedding & 29.4 & 22.3 & 48.1 & 43.1 & 143.0 & 64.9 & 64.1 & 47.5 & 47.4 \\ \hline
		\end{tabular}
	\end{adjustbox}
\end{table}
\subsection{Versatility}
To comprehensively evaluate the framework's flexibility in integrating different captioning models, we have implemented the Pointer Generation Network   \cite{131_see-etal-2017-get} (PGN), Single Layer RNN (SL-RNN) Network \cite{110_yao2015describing}, and Multi-Layer RNN (ML-RNN) network \cite{132_anderson2018bottom}. Except for the outlined distinctions, the remaining components are implemented identically without any alterations. As shown in Table \ref{table_versatlity}, incorporating these models leads to consistent enhancements across all emotional, factual, and holistic evaluation metrics. The successful integration of our model, SPECTRUM, with various captioning models demonstrates its adaptability, enhancing the emotionally-informed and field-acquainted performance of these models.
\begin{table}[!t]
	\centering
	\caption{Effect of Integrating Different Models with our Proposed Model on EMVIDCAP}
	\label{table_versatlity}
	\begin{adjustbox}{max width=\textwidth}
		\scriptsize
		\begin{tabular}{|>{\centering\arraybackslash}m{1.2cm}|>{\centering\arraybackslash}m{1.7cm}|>{\centering\arraybackslash}m{0.6cm}|>{\centering\arraybackslash}m{0.6cm}|>{\centering\arraybackslash}m{0.6cm}|>{\centering\arraybackslash}m{0.6cm}|>{\centering\arraybackslash}m{0.7cm}|>{\centering\arraybackslash}m{0.75cm}|>{\centering\arraybackslash}m{0.65cm}|>{\centering\arraybackslash}m{0.65cm}|>{\centering\arraybackslash}m{0.65cm}|}
			\hline
			\multicolumn{2}{|c|}{\multirow{3}{*}{{Model}}} &  \multicolumn{5}{c|}{\multirow{2}{*}{{Factual Metrics}}} &
			\multicolumn{2}{c|}{\parbox[c][2.0em][c]{1.2cm}{\centering \vspace{1.4em}{Emotional \\ Metrics}}} & \multicolumn{2}{c|}{\parbox[c][2.0em][c]{1.2cm}{\centering \vspace{1.4em}{Holistic \\ Metrics}}} \\  
			\multicolumn{2}{|c|}{} & \multicolumn{5}{c|}{} & \multicolumn{2}{c|}{} & \multicolumn{2}{c|}{}\\ \cline{3-11}
			\multicolumn{2}{|c|}{} & {B-4} & {M} & {R} & {C} & {Sum} & {Acc\textsubscript{sw}} & {Acc\textsubscript{c}} & {BFS} & {CFS} \\ \hline
			
					\multirow{2}{*}{PGN} & Base & 27.7 & 22.0 & 47.8 & 40.7 & 138.2 & 77.3 & 75.6 & 50.0 & 47.9 \\ \cline{2-11}
			& {SPECTRUM} + Base & \textbf{30.7} & \textbf{22.9} & \textbf{49.0} & \textbf{45.4} & \textbf{148.0} & \textbf{78.1} & \textbf{76.3} & \textbf{51.9} & \textbf{51.8} \\ \hline
			\multirow{2}{*}{SL-RNN} & Base & 25.3 & 20.0 & 46.8 & 34.7 & 126.9 & 62.5 & 60.7 & 46.7 & 40.1 \\ \cline{2-11}
			& {SPECTRUM} + Base & \textbf{26.0} & \textbf{20.9} & \textbf{47.7} & \textbf{37.1} & \textbf{131.7} & \textbf{64.5} & \textbf{62.8} & \textbf{46.9} & \textbf{42.4} \\ \hline
			\multirow{2}{*}{ML-RNN} & Base & 23.7 & 20.5 & 46.9 & 35.5 & 126.6 & 64.1 & 62.6 & 46.2 & 41.1 \\ \cline{2-11}
			& {SPECTRUM} + Base & \textbf{25.2} & \textbf{20.8} & \textbf{47.7} & \textbf{35.8} & \textbf{129.5} & \textbf{65.4} & \textbf{63.8} & \textbf{47.4} & \textbf{41.6} \\ \hline
		\end{tabular}
	\end{adjustbox}
\end{table}
\subsection{Qualitative Analysis}
\begin{figure*}[!t]
	\centering
\begin{subfigure}[b]{0.45\textwidth}
	\centering
	\includegraphics[height = 1.85cm, width=9.6cm]{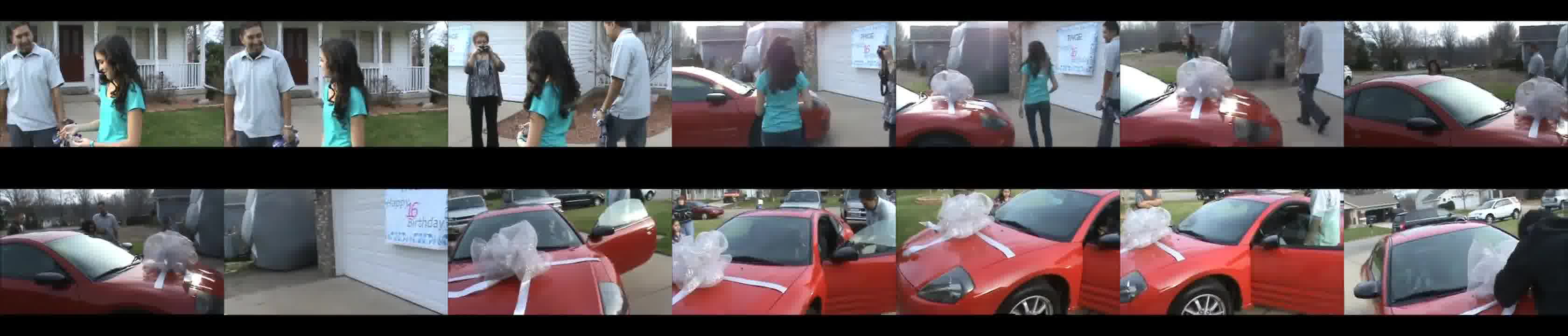}

	\vspace{5pt}  
	
	\begin{minipage}{1.1\textwidth} 
		\vspace{5pt} 
		\small 
		
		{\scriptsize \textbf{GT1}: a girl's father made a surprise which is a new car for her.}\\
		{\scriptsize \textbf{GT2}: the girl feels surprised at sight of the nice gift a new car.}\\
		{\scriptsize \textbf{Baseline}: two \textcolor{red}{worried} people are \textcolor{red}{talking} together \textcolor{red}{\textbf{calmly}}.}\\
		{\scriptsize \textbf{SPECTRUM}: the \textcolor{blue}{girl} was {surprised} \textcolor{blue}{when} \textcolor{blue}{she} \textcolor{blue}{saw} the \textcolor{blue}{new} \textcolor{blue}{car}.}
	\end{minipage}
	
	\vspace{5pt}  
	
	\includegraphics[height = 2.85cm, width=9cm]{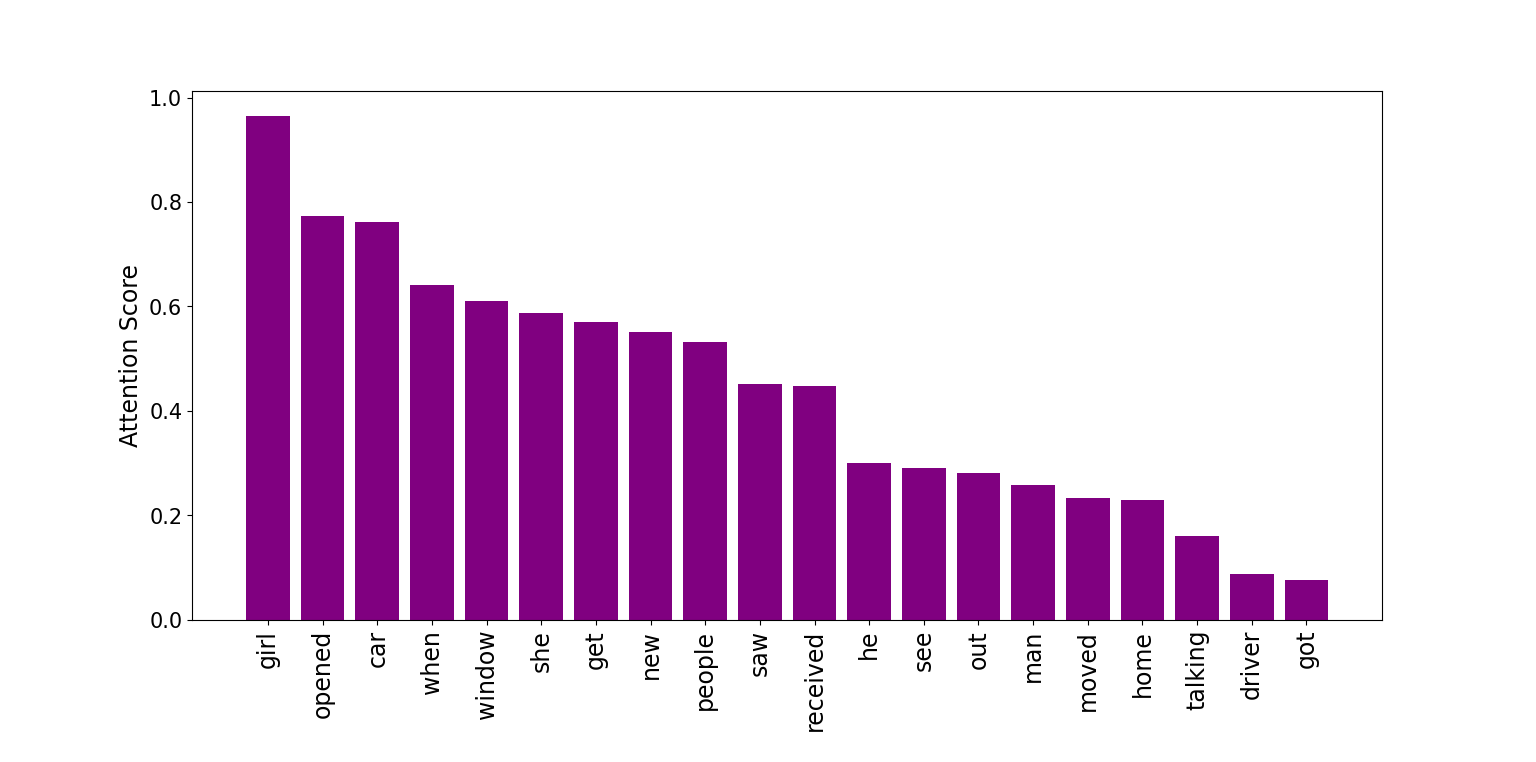}
	\caption{\footnotesize {EmVidCap Dataset. Video No. 70201.}}
	\label{subfig_qualitative_1804_red_car}
\end{subfigure}
	\hfill 
		\begin{subfigure}[b]{0.45\textwidth}
		\centering
		\includegraphics[height = 1.85cm, width=9.6cm]{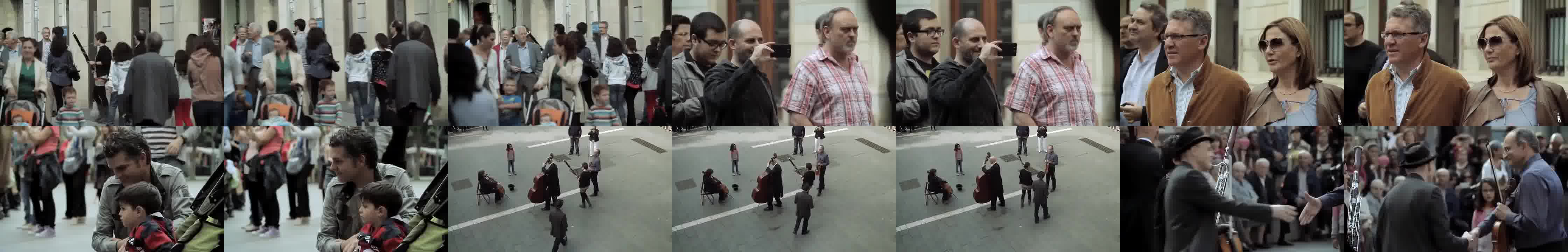}
		
		\vspace{5pt}  

		\begin{minipage}{1.1\textwidth} 
			\vspace{5pt} 
			\small 
			{\scriptsize \textbf{GT1}: the pleasant music from the cello attracts a lots people to listen.}\\
			{\scriptsize \textbf{GT2}: a band plays a happy song on the street and the people stop to watch.}\\
			{\scriptsize \textbf{Baseline}: people are playing music on the street \textcolor{red}{\textbf{surprisingly}}.}\\
			{\scriptsize \textbf{SPECTRUM}: a \textcolor{blue}{group} of \textcolor{blue}{people} are \textcolor{blue}{playing} the \textcolor{blue}{cello} {\textbf{happily}} in the \textcolor{blue}{street}.}
		\end{minipage}
		
		\vspace{5pt}  
		
		\includegraphics[height = 2.85cm, width=9cm]{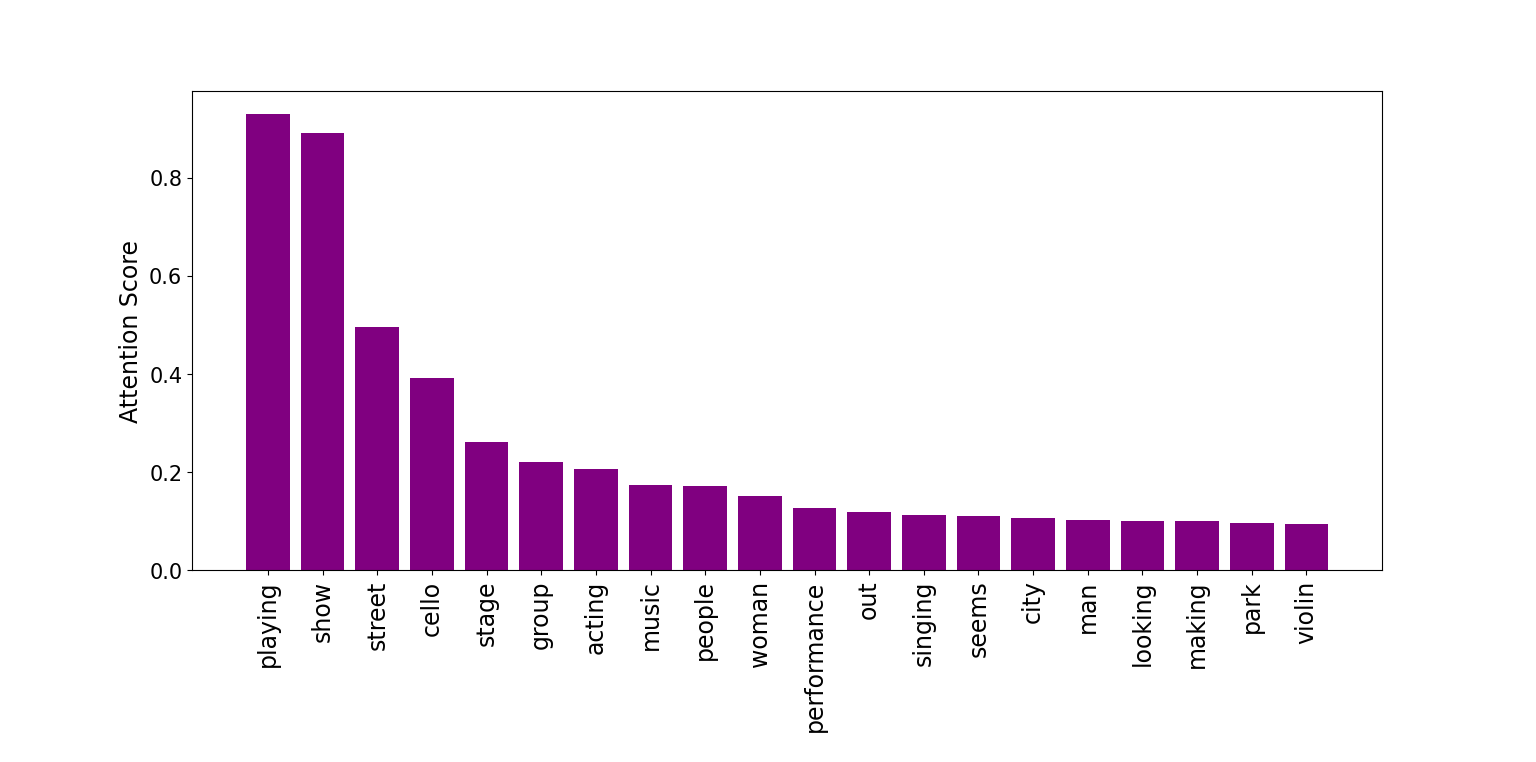}
		\caption{\footnotesize {EmVidCap Dataset. Video No. 51212.}}
		\label{subfig_qualitative_1728_cello_on_street}
	\end{subfigure}
	
	\caption{Captioning examples from EmVidCap Dataset, including two human-annotated ground-truth captions, captions generated by the Baseline and proposed SPECTRUM models, and bar charts showing the attention score of factual words over detected concepts. \textcolor{blue}{Accurate factual concepts}, and \textcolor{red}{wrongly predicted concepts}  are pointed out.}
\label{fig_qualitative_factual_captioning_bar_chart}
\end{figure*}

Fig \ref{fig_qualitative_factual_captioning_bar_chart} shows two caption examples with bar charts depicting attention scores for factual concepts. These examples highlight SPECTRUM's ability to produce field-acquainted captions that accurately reflect the factual context of the videos, demonstrating its effectiveness beyond just emotional detailing.
In Fig \ref{subfig_qualitative_1804_red_car} and Fig \ref{subfig_qualitative_1728_cello_on_street}, the critical factual concepts receive higher attention scores, which enhances the captions by focusing on more detailed aspects of the scenes. This underscores the premise that 'attention is all we need.' Our innovative approach to calculating attention using multimodal and attribute features leads to captions that closely align with these attributes, as shown in the bar charts. These charts demonstrate that the attention scores effectively represent the relationships between attribute and multimodal features, resulting in more accurate and detailed captions.\\
These six examples demonstrate the superiority of our model, SPECTRUM. The first aspect of this superiority stems from its ability to detect the predominant emotion of the video context. The second aspect is its ability to identify meaningful and relevant factual concepts, generating enriched, detailed captions.\\
\section{CONCLUSION}
 \noindent In this paper, we introduce SPECTRUM (Semantic Processing and Emotion-Informed video-Captioning Through Retrieval and Understanding Modalities), a novel framework designed to generate emotionally accurate and semantically resonant captions. SPECTRUM captures multimodal emotional and factual undertones by analyzing video-text attributes and utilizing a holistic, concept-oriented approach to determine the direction of emotionally-informed and field-acquainted captions. The framework incorporates estimating emotional probabilities, the adaptive weighting of embedded attribute vectors, and applying coarse- and fine-grained emotional concepts to ensure contextual alignment within video content. SPECTRUM leverages two loss functions to optimize the model’s performance to integrate emotional information while minimizing prediction errors. Extensive experiments conducted on the EmVidCap, MSVD, and MSR-VTT video captioning datasets, along with visualization results, demonstrate the state-of-the-art effectiveness of SPECTRUM in accurately capturing and conveying video emotions and multimodal attributes.

\bibliographystyle{IEEEtran}
\bibliography{ReferenceFile}		
		\end{document}